# Correlation over Decomposed Signals: A Non-Linear Approach to Fast and Effective Sequences Comparison

Luciano da Fontoura Costa

Cybernetic Vision Research Group

IFSC, University of São Paulo,

Caixa Postal 369

São Carlos, SP, Brazil

A novel non-linear approach to fast and effective comparison of sequences is presented, compared to the traditional cross-correlation operator, and illustrated with respect to DNA sequences.

***Introduction***: A new approach to the effective comparison of two sequences is introduced. The technique is based on a novel non-linear operator, based on the decomposition of the sequences and correlations performed in the Fourier domain [1], which is presented, compared with the traditional cross-correlation operator, and illustrated for identification of partial matches between two DNA sequences. Possible enhancements involving low-pass filtering of the original sequences are also commented and illustrated.

***The Comparison Operator:***   Let $s = [s_i]$ and $q = [q_i]$; $i = 1, 2, \cdots, N$; be one-dimensional sequences (or signals) discrete along their domains and amplitudes. In addition, let the amplitude of these sequences be limited and discrete, i.e. $s_i, q_i \in \{min, min+1, \cdots, max-1, max\}$, and define *L = max − min + 1*. The comparison between these two sequences involves the following three basic steps:

**(i)** *Sequence Normalization and Decomposition:* The two sequences are firstly proportionally normalized in such a way that $s_i, q_i \in \{1, 2, \cdots, M\}$, *M ≤ L*, and subsequently decomposed into respective binary signals $b^j = [b_i^j]$ and $c^j = [c_i^j]$ by making:

$$b_i^j = \begin{cases} 1 & if \ s_i = j \\ 0 & otherwise \end{cases}, \ for \ j = 1, 2, \cdots, M \qquad (1)$$

and

$$c_i^j = \begin{cases} 1 & if \ q_i = j \\ 0 & otherwise \end{cases}, \ for \ j = 1, 2, \cdots, M \qquad (2)$$

This decomposition is illustrated by the following example, assuming *N*=25 and *M*=2:

$s =$    2 1 2 2 2 1 1 1 1 2 1 2 1 1 2 2 2 2 1 1 2 1 2 1 1
$q =$    1 1 1 2 1 1 1 1 2 1 2 1 1 2 2 1 1 1 1 2 2 1 2 1 2

$b^1 =$   0 1 0 0 0 1 1 1 1 0 1 0 1 1 0 0 0 0 1 1 0 1 0 1 1
$c^1 =$   1 1 1 0 1 1 1 1 0 1 0 1 1 0 0 1 1 1 1 0 0 1 0 1 0

$b^2 =$   1 0 1 1 1 0 0 0 0 1 0 1 0 0 1 1 1 1 0 0 1 0 1 0 0
$c^2 =$   0 0 0 1 0 0 0 0 1 0 1 0 0 1 1 0 0 0 0 1 1 0 1 0 1

**(ii)** *Cross-Correlations*: Once the binary sequences are obtained, they are respectively correlated by pairs to yield $k_p^j = \sum_{i=1}^{N} b_i^j c_{i+p}^j$, $p = -N+1, \cdots, N-1$, which can be conveniently performed in the Fourier domain in $O(N \log N)$. Each correlation value indexed by $i$ in these correlation sequences corresponds precisely to the number of hits of the type "1 = 1" between the first binary sequence and the displaced version of the second, as illustrated below with respect to the above sequences:

$b_i^1 =$           0 1 0 0 0 1 1 1 1 0 1 0 1 1 0 0 0 0 1 1 0 1 0 1 1
$c_{i+3}^1 =$       1 1 1 0 1 1 1 1 0 1 0 1 1 0 0 1 1 1 1 0 0 1 0 1 0
$b_i^j c_{i+3}^j =$ 0 0 0 0 1 0 0 0 0 1 0 1 0 0 0 1 1 0 0 0 0 1 0 0 0 0 0
$k_p^j = \mathbf{6}$

**(iii)** *Adding the partial correlations*: The coincidence signal, the final product of the comparison operator, is obtained by adding all the above cross-correlations, i.e.:

*Correlation over decomposed signals..., L. da F. Costa, 7$^{th}$ May 2000*                    3

$$E_i = \sum_{j=1}^{M} k_i^j, \ i = -N+1, \ldots, N-1 \tag{2}$$

The obtained coincidence signal has the remarkable property that each of its values indicates the number of matches between elements of the first sequence and the respective displaced version of the second. More specifically, each peak in the coincidence signal indicates that the two sequences present several coincidences at the respective displacement. Moreover, the peak magnitude gives the respective number of hits. It should be also observed that the comparison operator is not affected by scaling by a same factor both the original sequences (non-linearity), which ultimately underlies its interesting potential for comparison.

***Results:*** The potential of the comparison operator can be better appreciated by comparing it with the traditional cross-correlation between two sequences. Figure 1 presents two DNA sequences to be compared (a) and (b), where the C, A, T, and G bases have been represented as 1, 2, 3, and 4, respectively, as well as the sequences obtained by applying the correlation (c) and comparison (d) operators. The first original sequence contains 512 elements and shares several pieces with the second sequence, especially the large block of 130 adjacent elements identified by an asterisk in (a). Actually, the second sequence is almost identical to the first, except for a few missing pieces, which represents to a particularly difficult situation for the comparison. A clearly superior result has been verified for the comparison operator (d), while the correlation (c) provided virtually no discriminating properties. The total execution time was less than 10ms in an IBM-PC 500MHz compatible personal microcomputer. Longer sequences involving 8192 bps can be processed in less than 50ms.

[Figure 1]

Since the matches involving short contiguous sequences or, even between isolated points, are also added to the overall result, a mean background "noise" of $N/4$, considering uniformly distributed random sequences, is obtained. This noise can be attenuated by convolving both the original sequences with rectangular functions of width $w$ and unit height (low-pass filtering). Figure 1(e) and (f) illustrates the effect of this strategy with respect to $w = 1.5$ and $4.5$, respectively.

**Concluding Remarks**: A novel non-linear fast and effective approach to the identification of partial (and also global) matches between two general sequences has been presented. Since it considers the identity of the sequence elements, and is not affected by their respective magnitudes, the obtained results correspond precisely to the number of matches between the sequences for the several relative displacements. The potential of the technique has been illustrated and compared to the traditional cross-correlation operator with respect to DNA sequences involving thousands of base pairs, corroborating the greater effectiveness of the comparison operator. Since the *M* involved cross-correlations can be effectively performed in the Fourier domain, the overall implied complexity is of $O(MN\text{Log}N)$. However, in the case where *N* is much larger than *M*, such as with DNA and RNA sequences, this complexity tends to $O(N\text{Log}N)$. A possibility for peak enhancement by convolving the original sequences with the rectangular function has also been presented and illustrated. The proposed technique represents great potential not only for gene and protein sequence analysis, but also for general and statistical signal and image recognition.

**Captions**

Figure 1 – Two original sequences (a-b), and the respective sequences obtained by applying the cross-correlation (c) and comparison (d) operators. The comparison obtained for the sequences convolved with rectangular boxes with *w* = 1.5 and 4.5 are shown in (e) and (f), respectively.

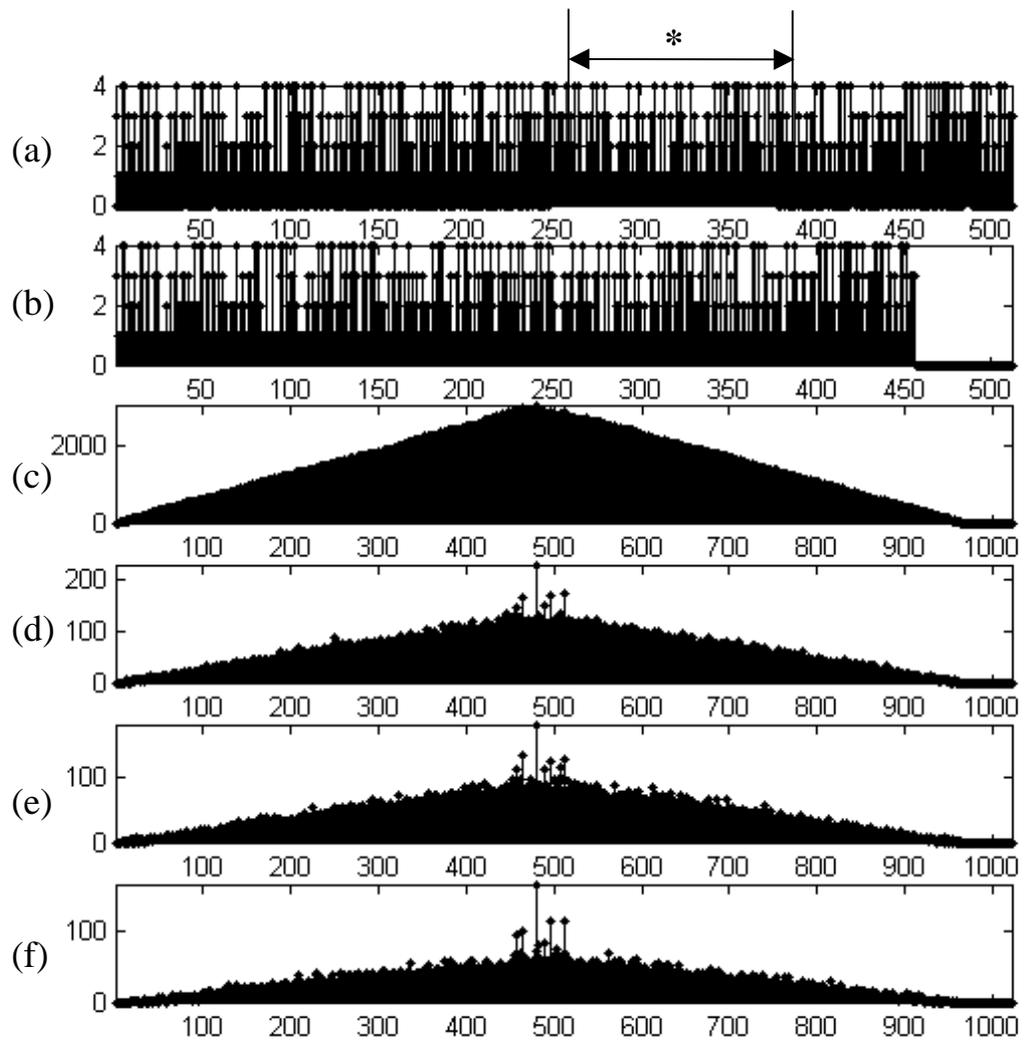